\documentclass[a4paper]{spie}  

\usepackage{amsmath,amsfonts,amssymb}
\usepackage{graphicx}
\usepackage[colorlinks=true, allcolors=blue]{hyperref}
\usepackage{geometry}
\usepackage[utf8]{inputenc}
\usepackage{float}
\usepackage{subcaption}

\title{Colony Grounded SAM2: Zero-shot detection and segmentation of bacterial colonies using foundation models}

\author{\text{Daan Korporaal}$^{*1}$, \text{Patrick de Kruijf}$^1$, \text{Ralph H.G.M. Litjens}$^1$, \text{Bas H.M. van der Velden}$^1$}
\affil{$^1$\text{Wageningen Food Safety Research, Wageningen, The Netherlands}}
\affil{$^*$\text{Corresponding author, email: daan.korporaal@wur.nl}}

\begin{document} 
\maketitle
\begin{abstract}
The detection and classification of bacterial colonies in images of agar-plates is important in microbiology, but is hindered by the lack of labeled datasets. Therefore, we propose Colony Grounded SAM2, a zero-shot inference pipeline to detect and segment bacterial colonies in multiple settings without any further training. By utilizing the pre-trained foundation models Grounding DINO and Segment Anything Model 2, fine-tuned to the microbiological domain, we developed a model that is robust to data changes. Results showed a mean Average Precision of 93.1\% and a $Dice@detection$ score of 0.85, showing excellent detection and segmentation capabilities on out-of-distribution datasets. The entire pipeline with model weights are shared open access to aid with annotation- and classification purposes in microbiology.
 
\end{abstract}


\section{Introduction}
\label{sec:intro}  
Bacterial colonies grown on agar plates allow quantification and identification for medical diagnostics, risk assessment, or food safety. Manual quantification is a complex and time-consuming task, sparking the need for automation. Recent advances in bacterial colony detection using artificial intelligence (AI) have shown impressive results\cite{wang2025colony,yang2023microbial,whipp2022yolo}. However, to perform well these AI models typically need task-specific manually labeled datasets, which is time- and resource-intensive. \\\\
To alleviate this need for manually labeled datasets, we propose Colony Grounded SAM2, a general purpose zero-shot detection and segmentation pipeline for bacterial colony images. By adapting large pre-trained foundation models to the microbiological domain, we aim for a robust method to detect and segment bacterial colonies over multiple datasets without needing further training. This allows researchers and domain experts to utilize AI-automated colony detection on data that was not feasible before, supporting and enhancing human observer performance in such tasks.

\section{Data}
We used three open access image datasets of colonies cultured on agar plates (\autoref{tab:datasets}). The AGAR (Annotated Germs for Automated Recognition) dataset \cite{majchrowska2021agar} consists of 5 types of pharmaceutical bacterial colonies on Trypticase Soy Agar plates, over multiple imaging settings. The ADBC (Annotated dataset for deep-learning-based bacterial colony detection) dataset \cite{makrai2023annotated} contains 24 bacteria species of veterinary importance on 4 agar types. Lastly, the Hemolysis segmentation dataset\cite{savardi2018automatic} contains annotated colony segmentation masks from blood agar plates, used to classify hemolysis producing bacterial colonies from urinary infections and throat swab screening.
\begin{table}[H]
\caption{AGAR = Annotated Germs for Automated Recognition, ADBC = Annotated Dataset for deep-learning-based Bacterial Colony detection.}
\label{tab:datasets}
\begin{center}
\begin{tabular}{l r r l }

\hline
\textbf{Dataset name}               & \textbf{\# photos}  & \textbf{\# colonies} & \textbf{Type of ground truth}   \\ 
\hline
AGAR\cite{majchrowska2021agar}& 18 000    & 336 442      &  Bounding boxes\\
ADBC\cite{makrai2023annotated}                                & 369       & 56 865       & Bounding boxes\\
Hemolysis segmentation\cite{savardi2018automatic}              & 286      & \textit{Unknown}        & Segmentations  \\
\hline

\end{tabular}
\end{center}
\end{table}
\section{Methods}
\label{sec:methods}
\begin{figure} [H]
   \begin{center}
   \includegraphics[width=\textwidth]{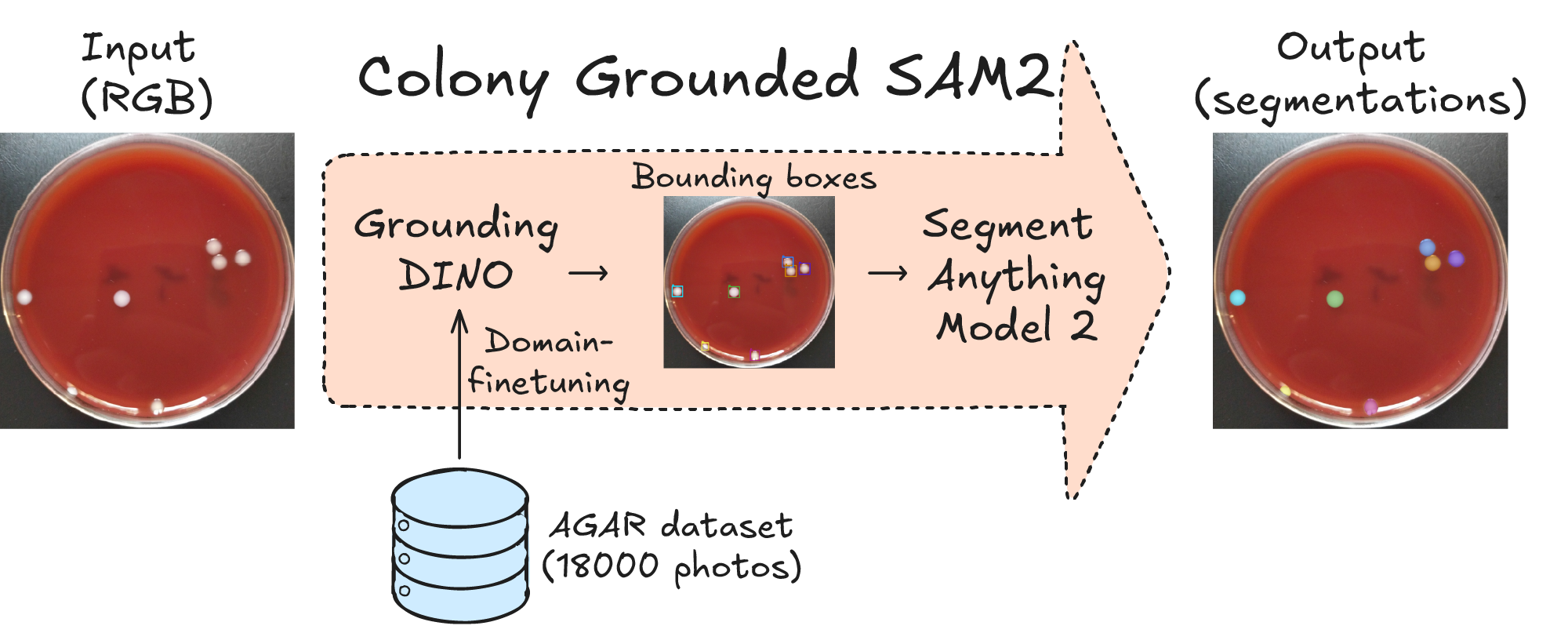}
   \end{center}
   \caption[example] 
   { \label{fig:example} 
The proposed Colony Grounded SAM2 pipeline for bacterial colonies}
\end{figure}

We propose Colony Grounded SAM2 (\autoref{fig:example}), an AI pipeline for zero-shot inference that consists of sequential detection and segmentation of bacterial colonies. We used Grounded SAM2\cite{ren2024grounded}, a combination of Grounding DINO\cite{liu2023grounding} and Segment Anything Model (SAM) 2\cite{ravi2024sam2segmentimages}, and fine-tuned the Grounding DINO to a bacterial colony database. Code and models are available at \url{https://github.com/WFSRDataScience/ColonyGroundedSam2}. The following subsections further explain these models and our evaluation of the zero-shot AI pipeline.
\subsection{Detection}
The first task of our pipeline was to detect individual colonies in the images using Grounding DINO. Grounding DINO extends the large pre-trained vision transformer DINO\cite{caron2021emerging} to an open-vocabulary detection model using grounded pre-training. The open-vocabulary and self-supervised pre-trained vision transformer give Grounding DINO a broad and robust understanding of object detection in any natural image, making Grounding DINO a good choice for zero-shot approaches. \\\\
Microbiological agar plate images are, however, very different from Grounding DINO's pre-training with natural images, making its out-of-the-box performance on such images subpar. Therefore, we decided to adapt Grounding DINO to this unseen image domain with brief domain-specific fine-tuning using the AGAR dataset\cite{majchrowska2021agar}. To keep the inherent robustness and zero-shot capability of Grounding DINO, we chose to fine-tune for only a single epoch using the recommended setting\cite{liu2023grounding}.\\\\
We evaluated the fine-tuned model using the ADBC dataset\cite{makrai2023annotated} and compared to the five baseline models from Colony-YOLO\cite{wang2025colony}. Since the ADBC dataset includes many more types of microorganisms on different agar plates (\textit{e.g.} blood and chocolate blood agar), this dataset allows for a broad evaluation of the zero-shot capabilities of our adapted model.  We evaluated our model on the mean Average Precision (mAP). Matches were determined using the Intersection-over-Union (IoU). Because some annotated boxes were found to be loosely or imprecisely labeled (see \autoref{fig:qual4}), we chose an IoU threshold of 0.2. This value was chosen empirically to ensure that no incorrect matches were introduced, while minimizing the penalty on our model for predicting precise bounding boxes around relatively small objects.
\subsection{Segmentation}
The bounding boxes from our detection model were fed directly to Segment Anything Model 2 (SAM2), a large-scale pre-trained foundation model, designed to segment any object in an image\cite{ravi2024sam2segmentimages}. SAM2 can take both the image and the found bounding boxes as a prompt as input, and return the instance segmentation mask of each found colony. We hypothesized that since we prompt with the bounding boxes from Grounding DINO, SAM2 is already instructed where to focus its attention and thus further domain adaptation was unnecessary.\\\\
We evaluated the segmentations using the Hemolysis segmentation dataset\cite{savardi2018automatic} using the Dice score. Additionally, to assess the performance of the segmentation regardless of detection performance, we calculated the $Dice@detection$ score, defined as the Dice score on detected areas by Grounding DINO. While the regular Dice score thus describes the overall performance of our pipeline, the $Dice@detection$ score describes the SAM2 segmentation performance with the given bounding boxes.
\section{Results}
\label{sec:results}
\subsection{Detection}
\begin{figure}[H]
    \centering

    \begin{subfigure}[b]{0.4\textwidth}
        \includegraphics[width=\textwidth, keepaspectratio]{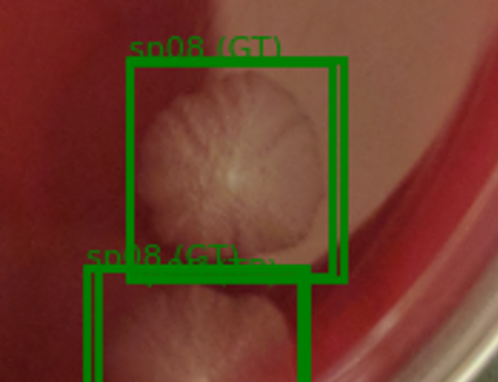}
        \caption{Good generalization on out-of-distribution colony types}
        \label{fig:qual1}
    \end{subfigure}
    \hfill
    \begin{subfigure}[b]{0.4\textwidth}
        \includegraphics[width=\textwidth, keepaspectratio]{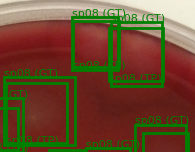}
        \caption{Good performance on hard-to-see objects}
        \label{fig:qual2}
    \end{subfigure}
    
    \begin{subfigure}[b]{0.4\textwidth}
        \includegraphics[width=\textwidth, keepaspectratio]{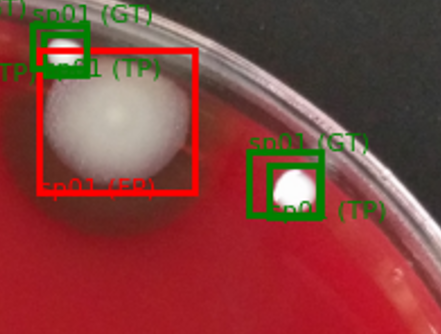}
        \caption{The model finds colonies of any other bacterial types than the focus of the plate}
        \label{fig:qual3}
    \end{subfigure}
    \hfill
    \begin{subfigure}[b]{0.4\textwidth}
        \includegraphics[width=\textwidth, keepaspectratio]{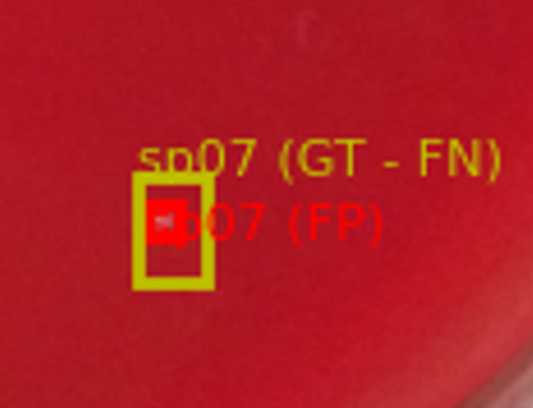}
        \caption{The model detects small objects very precisely, leading to incorrect mismatches}
        \label{fig:qual4}
    \end{subfigure}
    \vspace{0.7em}
    \caption{Qualitative detection results illustrating model behavior across various conditions. Bacterial colony types are labeled by their ID as defined in Table 1 from the ADBC paper\cite{makrai2023annotated}. Both ground truths (GT) and predictions are shown. Predictions correctly matched to ground truths are green, unmatched ground truths are yellow (FN = False Negative), and unmatched predictions are red (FP = False Positive).}
    \label{fig:qualitative-examples}
\end{figure}

Our Colony Grounded SAM2 is on par or slightly outperforms the five baseline methods from Colony-YOLO in detection on the ADBC dataset (\autoref{tab:det_map}). Note that for these five baseline models, a positively predicted data point has to be both detected precisely and classified correctly, while our current implementation foregoes this classification step. Still, as Colony-YOLO tends to make little classification errors, the performance achieved by our zero-shot model shows commendable detection capabilities in comparison.
\begin{table}[H]
\caption{Detection performances on the ADBC dataset}
\label{tab:det_map}
\begin{center}
\begin{tabular}{ll}
\hline
Model                 & mAP/\% \\ \hline
Faster R-CNN          & 76.5   \\
YOLOv5n               & 82.6   \\
YOLOv8n               & 86.7   \\
YOLOv10n              & 87.4   \\
Colony-YOLO           & 91.1   \\ \hline
Colony Grounded SAM2 & 93.1*  \\ \hline
\end{tabular}
\caption*{\small \it{*As a set of binary classification tasks} }
\end{center}
\end{table}
Qualitative examples show that the model performs well in unseen or difficult circumstances \autoref{fig:qualitative-examples}. Furthermore, some of the mistakes found by the evaluation are understandable in the context of zero-shot inference, such as detecting colonies that were not the focus of that plate (\autoref{fig:qual3}) or very loose ground truths leading to wrongly counted negatives (\autoref{fig:qual4}). Over the entire dataset, Colony Grounded SAM2 seems to draw its bounding boxes much more precise than the annotated bounding boxes, especially on small colonies. This further highlights its strong capabilities as a pre-annotator for such datasets.
\subsection{Segmentation}
\begin{figure}[H]
    \centering

    \begin{subfigure}[b]{0.4\textwidth}
        \includegraphics[width=\textwidth, keepaspectratio]{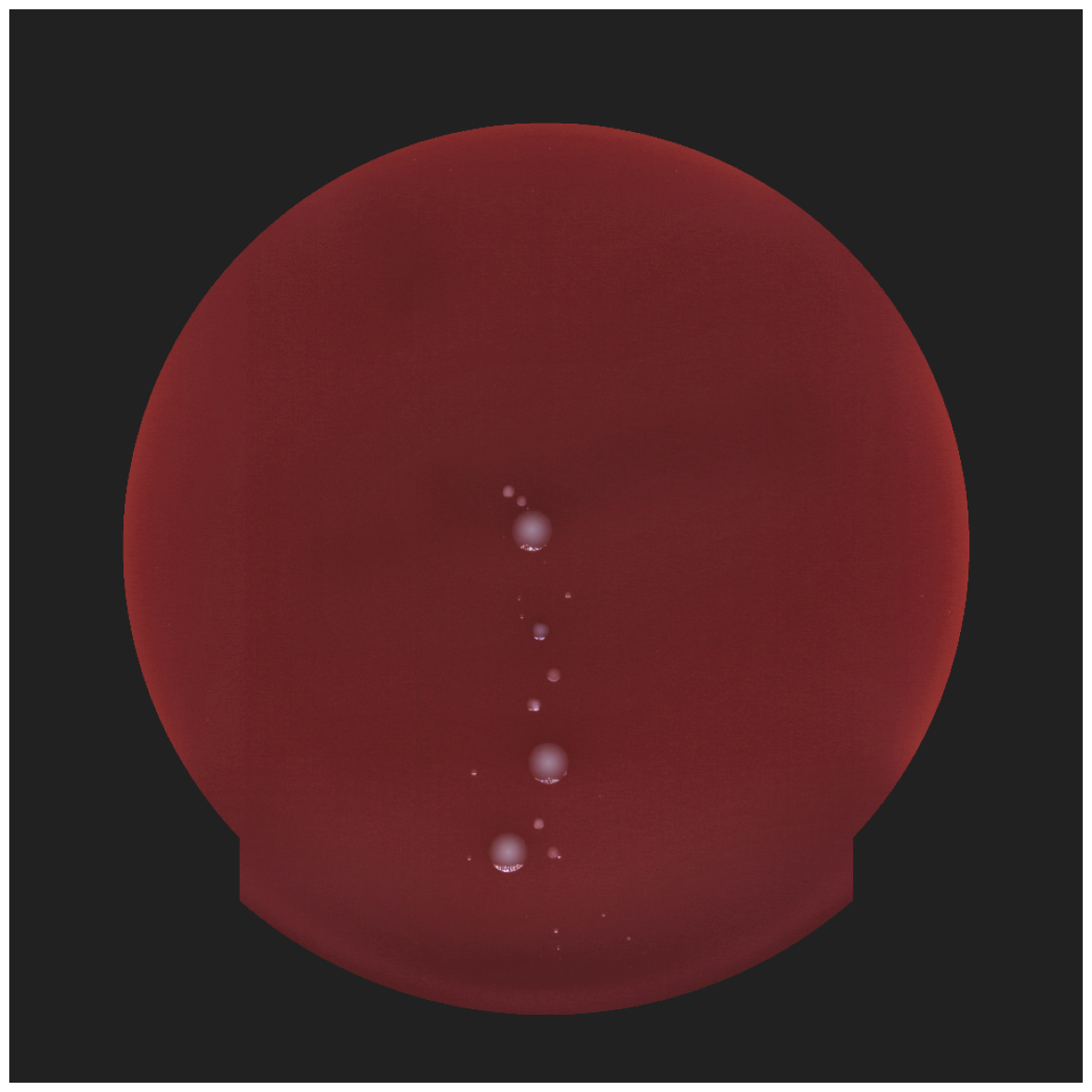}
        \caption{Example 1}
        \label{fig:ex1}
    \end{subfigure}
    \hfill
    \begin{subfigure}[b]{0.4\textwidth}
        \includegraphics[width=\textwidth, keepaspectratio]{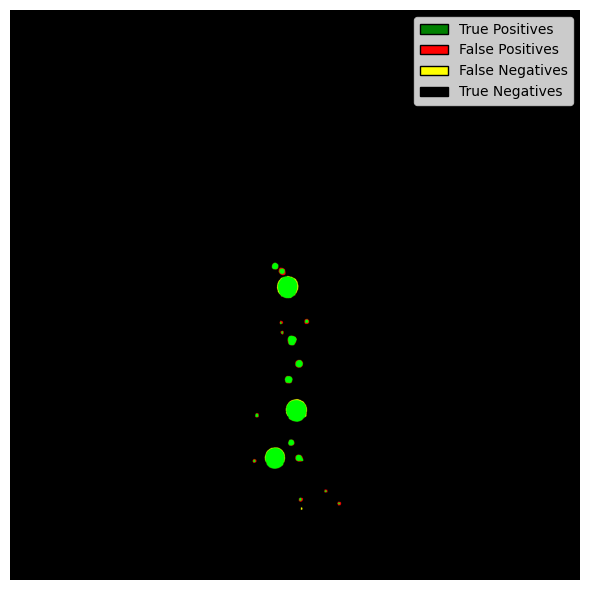}
        \caption{Result 1}
        \label{fig:res1}
    \end{subfigure}
    \\
    \begin{subfigure}[b]{0.4\textwidth}
        \includegraphics[width=\textwidth, keepaspectratio]{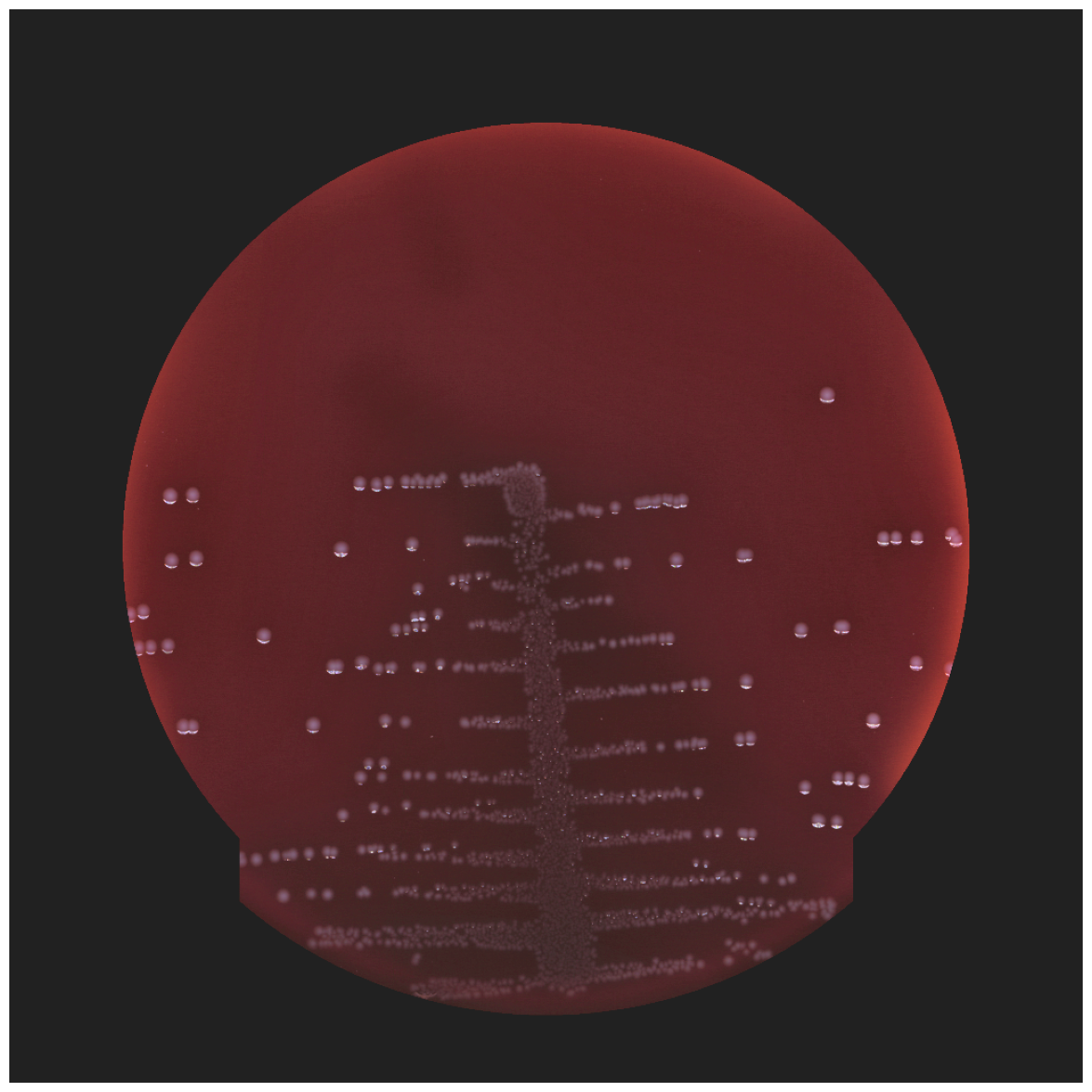}
        \caption{Example 2}
        \label{fig:ex2}
    \end{subfigure}
    \hfill
    \begin{subfigure}[b]{0.4\textwidth}
        \includegraphics[width=\textwidth, keepaspectratio]{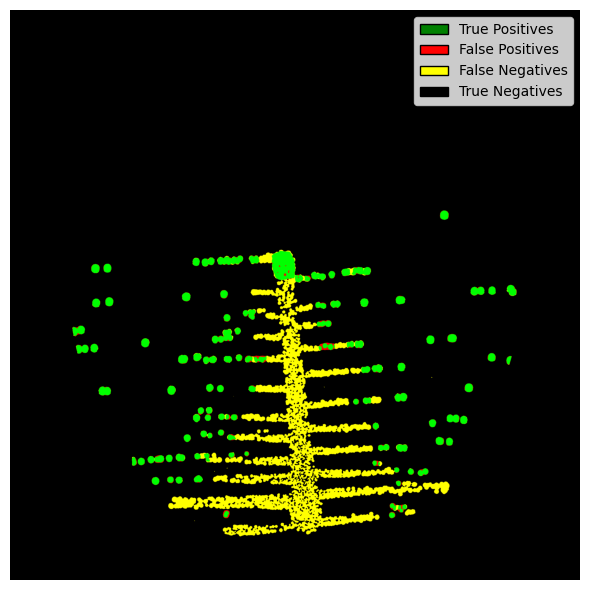}
        \caption{Result 2}
        \label{fig:res2}
    \end{subfigure}
    \vspace{0.7em}
    \caption{Qualitative segmentation results on the Hemolysis dataset. Correctly matched colonies are shown as green, unmatched ground truths are yellow, and unmatched predictions are red.}
    \label{fig:segmentation-rows}
\end{figure}

Overall segmentation performance on the Hemolysis segmentation dataset was 0.60 (Dice score). Additionally, the $Dice@detection$ score was 0.85, implying that our detection model made mistakes. This $Dice@detection$ score of 0.85 shows that our pure segmentation performance is high when compared to previous work\cite{cicatka2024increasing}. Additionally, Dice scores are known to be especially punishing towards segmenting smaller objects, as a few pixels of difference make a much larger impact in these cases\cite{maier2024metrics}. Therefore, in segmentation tasks such as provided here a $Dice@detection$ score of 0.85 usually corresponds to a very accurate performance. Qualitative analysis shows high quality segmentation masks, with only the smallest ground truth components showing slight differences in their segmentations  (\autoref{fig:res1}).  \autoref{fig:res2} shows high-quality segmentations, although many densely packed colonies in the center are missed. This pattern exists throughout the dataset, partly explaining differences in Dice and $Dice@detection$ scores. Since these central colonies are hard to distinguish and experts typically focus on more separable outer colonies, the model’s behavior is acceptable in practice.
\section{Conclusions}
We presented Colony Grounded SAM2, a zero-shot pipeline for the detection and segmentation of bacterial colonies. Utilizing a domain-adapted Grounding DINO detector, combined with a SAM2 segmenter, we have designed a highly robust, complex, yet fast model pipeline. With a mAP percentage of 93.1, and a $Dice@detection$ of 0.85, we achieved impressive performance on unseen datasets, highlighting the zero-shot capabilities of this pipeline on a wide variety of colony types, agar types, and image qualities. Colony Grounded SAM2 can be used for pre-annotation or classification of colonies on many different datasets without further modifications, further supporting the work of human experts.
\section*{Acknowledgements}
This work was supported by the Dutch Ministry of Agriculture, Fisheries, Food Security and Nature through the Food Safety Knowledge development program (project KB-56-002-006).\\
This research was funded by the Dutch Ministry of Agriculture, Fisheries, Food Security and Nature under the Wageningen University \& Research Knowledge Base Programme on Future Food Systems (KB-53-000-010). 
\bibliography{report} 
\bibliographystyle{spiebib} 

\end{document}